\documentclass[10pt,twocolumn,letterpaper]{article}

\usepackage{cvpr}              

\usepackage{multirow}
\usepackage{multicol}
\usepackage{makecell}
\usepackage{graphicx}
\usepackage{booktabs}
\usepackage{colortbl}
\usepackage{amsmath}
\usepackage{amssymb}
\usepackage[table]{xcolor} 
\usepackage{bbding}
\usepackage{tablefootnote}
\usepackage{titletoc}

\definecolor{cvprblue}{rgb}{0.21,0.49,0.74}
\usepackage[pagebackref,breaklinks,colorlinks,allcolors=cvprblue]{hyperref}
\def\paperID{1946} 
\def\confName{CVPR}
\def\confYear{2026}

\title{OmniVTG: A Large-Scale Dataset and Training Paradigm for \\ Open-World Video Temporal Grounding}

\author{Minghang Zheng$^1$ \quad Zihao Yin$^3$ \quad Yi Yang$^3$ \quad Yuxin Peng$^1$ \quad Yang Liu$^{1,2,4}$\thanks{Corresponding author}\\
$^1$Wangxuan Institute of Computer Technology, Peking University\\
$^2$State Key Laboratory of General Artificial Intelligence, Peking University\\
$^3$Central Media Technology Institute, Huawei Technologies Ltd.\\
$^4$PKU-WUHAN Institute for Artificial Intelligence, Peking University\\
{\tt\small \{minghang,pengyuxin,yangliu\}@pku.edu.cn} \quad
{\tt\small \{yinzihao6,yangyi16\}@huawei.com}
}

\begin{document}
\maketitle
\begin{abstract}

Video Temporal Grounding (VTG), the task of localizing video segments from text queries, struggles in open-world settings due to limited dataset scale and semantic diversity, causing performance gaps between common and rare concepts. To overcome these limitations, we introduce OmniVTG, a new large-scale dataset for open-world VTG, coupled with a Self-Correction Chain-of-Thought (CoT) training paradigm designed to enhance the grounding capabilities of Multimodal Large Language Models (MLLMs). Our OmniVTG is constructed via a novel Semantic Coverage Iterative Expansion pipeline, which first identifies gaps in the vocabulary of existing datasets and collects videos that are highly likely to contain these target concepts. For high-quality annotation, we leverage the insight that modern MLLMs excel at dense captioning more than direct grounding and design a caption-centric data engine to prompt MLLMs to generate dense, timestamped descriptions. Beyond the dataset, we observe that simple supervised finetuning (SFT) is insufficient, as a performance gap between rare and common concepts still persists. We find that MLLMs' video understanding ability significantly surpasses their direct grounding ability. Based on this, we propose a Self-Correction Chain-of-Thought (CoT) training paradigm. We train the MLLM to first predict, then use its understanding capabilities to reflect on and refine its own predictions. This capability is instilled via a three-stage pipeline of SFT, CoT finetuning, and reinforcement learning. Extensive experiments show our approach not only excels at open-world grounding in our OmniVTG dataset but also achieves state-of-the-art zero-shot performance on four existing VTG benchmarks. Code is available at \url{https://github.com/oceanflowlab/OmniVTG}.

\end{abstract}

\section{Introduction}
\label{sec:intro}

\begin{figure}[t]
    \centering
    \includegraphics[width=\linewidth]{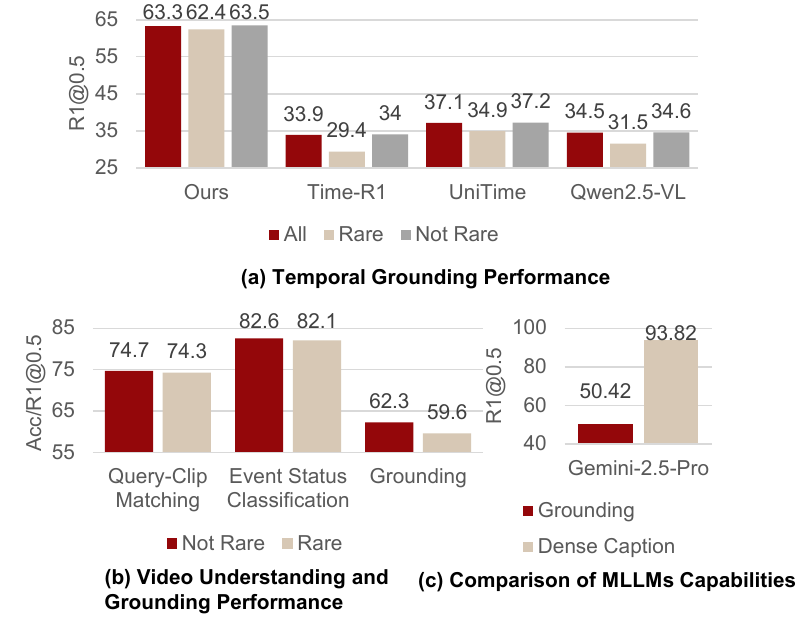}
    \caption{(a) Open-world video temporal grounding performance on our OmniVTG dataset. (b) The MLLM's (Qwen2.5-VL-7B~\cite{bai2025qwen25vltechnicalreport}) ability in video understanding and temporal grounding tasks. (c) The accuracy of timestamps in grounding and dense caption tasks performed by Gemini-2.5-Pro~\cite{comanici2025gemini25pushingfrontier}.}
    \label{fig:teaser_cmp}
\end{figure}

\begin{figure*}[t]
    \centering
    \includegraphics[width=\linewidth]{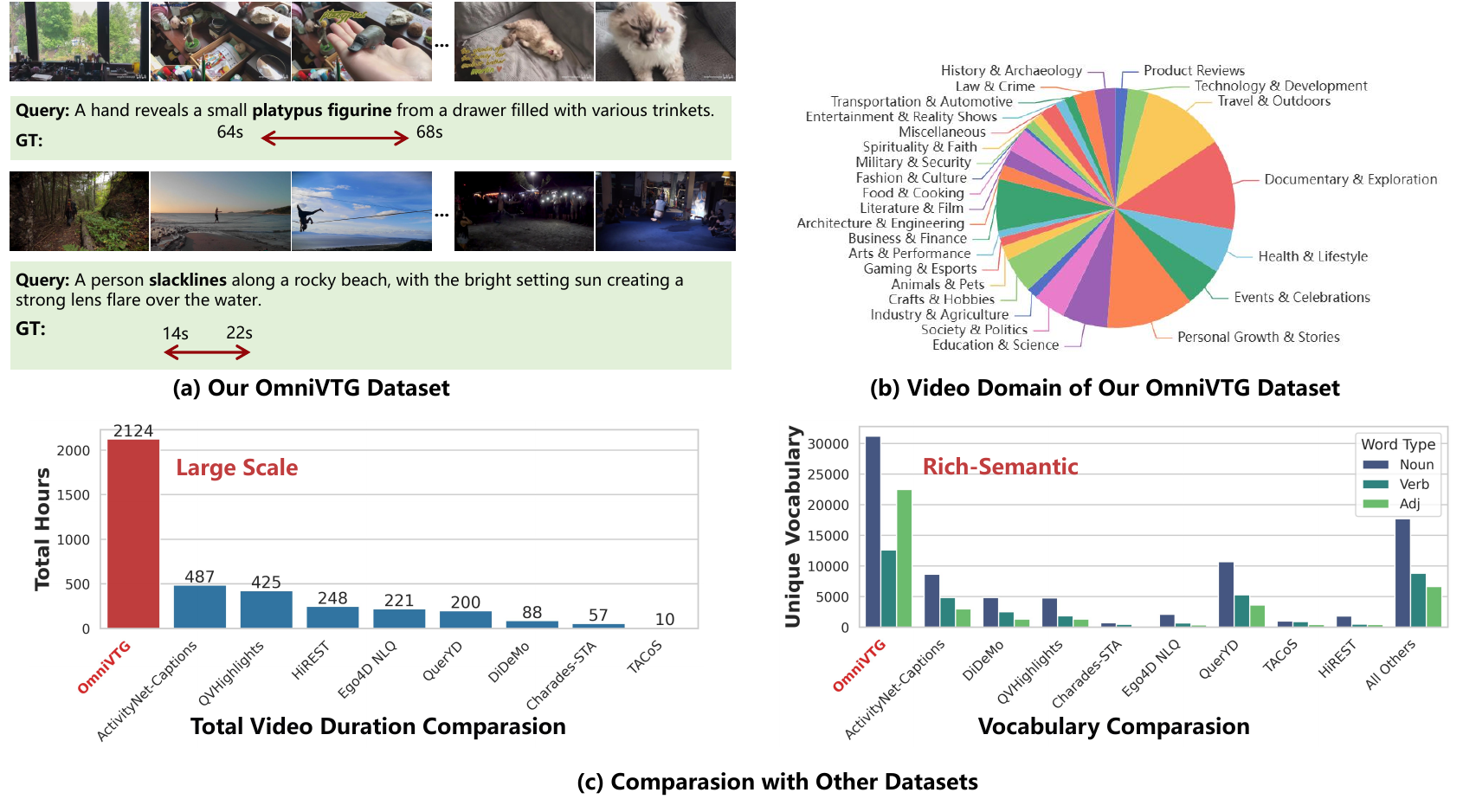}
    \caption{The visualizations and comparisons of our OmniVTG dataset.}
    \label{fig:teaser}
\end{figure*}

Video Temporal Grounding (VTG) is the task of identifying the precise start and end timestamps of events described by a natural language query within an untrimmed video. With the development of deep learning~\cite{bai2025qwen25vltechnicalreport,unitime2025,wang2025timer1,E250297,EN20140218} and big data~\cite{krishna2017dense,gao2017tall,yang20243d,mo2025advancing}, this task is being increasing important for real-world applications, such as video search and automated content analysis. 

In open-world scenarios, the challenge of VTG becomes significantly more complex. Real-world videos often contain a broad spectrum of events and concepts, ranging from common and everyday occurrences to rare, abstract, or even highly domain-specific phenomena. To effectively handle these diverse events, models must be able to understand a wide range of semantic concepts, including not only frequent and well-defined categories but also rare and ambiguous ones. As shown in Fig.~\ref{fig:teaser_cmp} (a), we find that the performance of existing Video Temporal Grounding MLLMs~\cite{unitime2025,wang2025timer1,bai2025qwen25vltechnicalreport} is significantly lower when handling these open-world scenarios, especially when they contain rare concepts. This reflects the limitations of existing datasets and models.

In terms of datasets, existing public VTG datasets fall short of the open-world requirements, suffering from two primary limitations. \textit{First, limited semantic coverage.} Mainstream datasets are often restricted to specific domains (e.g., indoor activities in Charades-STA~\cite{gao2017tall}) or possess a narrow vocabulary as shown in Fig.~\ref{fig:teaser} (c). \textit{Second, limited scale and quality.} Datasets relying on manual annotation~\cite{krishna2017dense,gao2017tall,lei2021detecting} are expensive and laborious to create, making them difficult to scale. Automated pipelines~\cite{soldan2022mad,oncescu2021querydvideodatasethighquality} often depend on sources like Automatic Speech Recognition (ASR), which cannot guarantee a precise match between the query (speech) and the visual content.
To address these data limitations, we introduce OmniVTG, a large-scale, semantic-rich open-world dataset designed to expand semantic coverage. Our collection process employs a \textit{Semantic Coverage Iterative Expansion} strategy, actively identifying underrepresented concepts and collecting relevant videos from the web. We begin by analyzing the vocabulary coverage of existing datasets to identify underrepresented rare concepts. We then perform targeted expansion by retrieving videos that are more likely to contain these rare concepts. To efficiently gather relevant videos, we leverage LLMs to translate target concepts into effective search keywords for video collection from the internet. To achieve automated annotation, we observed that the accuracy of timestamps generated during dense captioning is significantly higher than that obtained through direct grounding for modern MLLM~\cite{comanici2025gemini25pushingfrontier} as shown in Fig.~\ref{fig:teaser_cmp} (c)~\footnote{We use Gemini-2.5-Pro to generate dense captions and manually check the timestamps to report accuracy. Then, we use the same model to ground this caption, and the grounding performance is much lower.}.
Based on this insight, we designed a caption-centric data engine. We prompt MLLM~\cite{comanici2025gemini25pushingfrontier} to generate dense, timestamped captions that explicitly cover our target rare concepts, ensuring high-quality and fully automated temporal annotations. As shown in Fig.~\ref{fig:teaser} (a) and (c), in this way, our dataset successfully covers rare vocabulary and enriches the query semantics.
Finally, we manually inspect and refine a subset of these annotations to create a robust test set for evaluating open-world grounding performance.

In terms of models, while Supervised Fine-Tuning (SFT) on OmniVTG improves overall performance, we still observe that the gap between rare and common concepts persists, as shown in Fig~\ref{fig:teaser_cmp}(b). This suggests the model needs a more robust reasoning mechanism to handle unfamiliar, rare concepts. Furthermore, we find that the model's capabilities in video understanding are significantly stronger than its direct grounding ability, and the performance gap between rare and common concepts is much smaller. For example, as shown in Fig.~\ref{fig:teaser_cmp} (b), the model is better at judging whether a given video segment matches a text query and judging the state of the event described by the query at a specific timestamp (i.e., not started, ongoing, or ended)~\footnote{We finetune Qwen2.5-VL on all tasks and test the performance.}. Based on this observation, we propose a \textit{Self-Correction Chain-of-Thought (CoT)}. We require the model to first make predictions and then reflect on them using its video understanding capabilities. For example, it should determine whether the predicted segment indeed matches the query. Additionally, it should assess the state of the predicted start and end time events, such as inferring that the start time needs to be moved backward if the event has not yet begun at the predicted start time. To achieve this, we construct CoT fine-tuning data using our OmniVTG dataset that embodies this self-reflection mechanism to train the model and then further enhance the model's reasoning abilities through reinforcement learning.

Our contributions are summarized as follows: (1) We introduce OmniVTG, a large-scale, semantic-rich dataset constructed via a novel Semantic Coverage Iterative Expansion strategy. (2) We propose a Self-Correction Chain-of-Thought mechanism by leveraging the model's video understanding performance to reflect and refine its temporal grounding predictions. (3) Comprehensive experiments show our approach not only excels at our manually annotated OmniVTG test set but also achieves state-of-the-art zero-shot performance on four public VTG benchmarks.

\section{Related Work}
\label{sec:relatedwork}

\subsection{Datasets for Video Temporal Grounding}

\textit{Data collection methods.} Prominent datasets~\cite{krishna2017dense,gao2017tall,lei2021detecting}, rely on intensive manual annotation. While this process yields high-quality labels, it is expensive and difficult to scale, resulting in a limited total data volume. In contrast, automated pipelines~\cite{oncescu2021querydvideodatasethighquality,miech2019howto100m,soldan2022mad} usually leverage sources like Automatic Speech Recognition (ASR) to create query-moment pairs. This approach is more scalable but cannot guarantee that the spoken query precisely aligns with the visual content. \textit{Domain coverage.} These datasets suffer from limited video domain and semantic coverage. Mainstream benchmarks like ActivityNet Captions~\cite{krishna2017dense}, Charades-STA~\cite{gao2017tall}, and TACoS~\cite{regneri2013tacos} primarily focus on common human activities. QVHighlights~\cite{lei2021detecting} is restricted to specific domains like vlogs and news. While open-domain datasets like QuerYD~\cite{oncescu2021querydvideodatasethighquality} exist, their vocabulary coverage remains insufficient as shown in Fig~\ref{fig:teaser} (c), failing to represent the rare real-world concepts. To address these problems, we propose a novel \textit{Semantic Coverage Iterative Expansion pipeline} and create a large-scale OmniVTG dataset. This pipeline iteratively discovers uncovered rare words and collects videos in a targeted manner, which significantly expands the scale of data and semantic coverage.

\subsection{Methods for Video Temporal Grounding}
Early task-specific VTG methods~\cite{gao2017tall,zhang2020learning,wang2022negative,li2024momentdiff,mun2020local,liu2024towards,xiao2024bridging,mdetr,soldan2022mad,hou2022cone,barrios2023guidance,pan2023scanning,mu2024snag,zheng2025weakly,zheng2025hierarchical} extracted video and text features using pre-trained encoders and then applied complex cross-modal fusion and temporal localization modules to predict event boundaries. However, these methods heavily rely on training on closed-domain datasets, limiting their zero-shot capabilities and hindering generalization to open-world scenarios. With the rise of Multi-modal Large Models (MLLMs)~\cite{bai2025qwen25vltechnicalreport,wang2024videollamb,E250336,E230263,yang2025ar,yang2025planllm}, recent methods have leveraged their superior multimodal comprehension and reasoning abilities~\cite{unitime2025,wang2025timer1,zeng2025distime,zheng2024training}. Some methods focus on time representation. For example, TimeChat~\cite{ren2024timechat} and UniTime~\cite{unitime2025} explicitly encode timestamps as text, TRACE~\cite{guo2024trace} and DisTime~\cite{zeng2025distime}  introduce additional encoders and decoders for time.  
Another research direction is the training paradigm. Many methods~\cite{huang2024vtimellm,unitime2025,ren2024timechat,guo2024trace,zengtimesuite,zeng2025distime} adopt multi-stage Supervised Finetuning (SFT) to enhance models' temporal localization abilities. More recently, Reinforcement Learning (RL) has been applied to optimize temporal reasoning ability. For instance, Time-R1~\cite{wang2025timer1} uses RL to guide the model in generating a thinking process before giving the answer. However, we observe that its generated thoughts often lack explicit reflection and correction processes, focusing mainly on describing video content or repeating the query. Despite rapid development, we find that these methods still struggle with rare video concepts in open-world settings. Our method addresses these gaps by proposing a Self-Correction CoT training paradigm, which explicitly leverages the model's understanding abilities to generate a self-correction reasoning process.

\begin{figure*}
    \centering
    \includegraphics[width=0.8\linewidth]{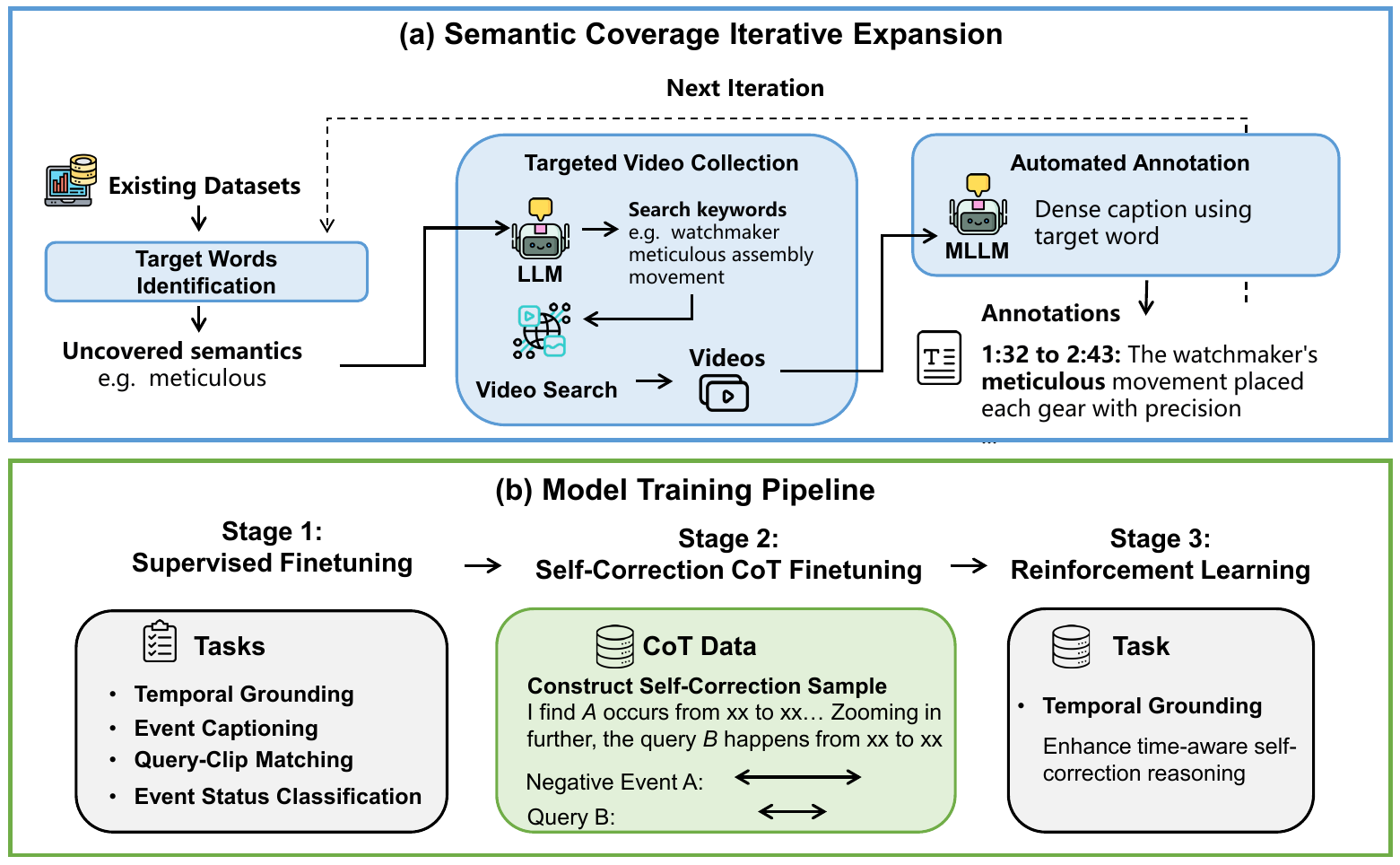}
    \caption{\textbf{(a) Our dataset collection pipeline.} The Target Words Identification identifies underrepresented words in existing datasets. The Interactive Video Collection collects videos that are more likely to contain the target word. The Automated Annotation reformulates the grounding tasks to dense caption tasks and prompts MLLMs to generate timestamps and captions using the target word. \textbf{(b) Our model training pipeline.} The Supervised Finetuning stage enhances basic temporal localization and the specific understanding skills needed for self-correction. The Self-Correction CoT Finetuning stage explicitly teaches the model the `predict-correct' reasoning path. The Reinforcement Learning stage further enhances the model's complex reasoning ability.
 }
    \label{fig:pipeline}
\end{figure*}

\begin{table*}[t]
    \centering
    \scalebox{0.8}{
    \begin{tabular}{l|ccc|cc|ccc|c}
    \toprule
    Dataset & \makecell{Total\\Duration} & \makecell{Duration \\ / Video} & \makecell{Duration \\ / Moment}  & \makecell{Total \\ Queries} & \makecell{\# Words\\/ Query} & Nones & Verbs & Adj. & Domain \\
    \midrule
    Anet-Captions~\cite{krishna2017dense} & 487 h & 1.96 min & 37.1 s & 72.0 K & 14.8 & 8.6 K & 4.8 K & 3.0 K & Activity \\
    Charades-STA~\cite{gao2017tall} & 57.1 h & 0.50 min & 8.1 s & 16.1 K & 7.2 & 0.7 K & 0.5 K & 0.2 K & Activity \\
    TACoS~\cite{regneri2013tacos} & 10.1 h & 4.78 min & 27.9 s & 18.2 K & 10.5 & 1.0 K & 0.9 K & 0.4 K & Cooking\\
    DiDeMo~\cite{didemo} & 88.7 h & 0.50 min & 6.5 s & 41.2 K & 8.0 & 4.8 K & 2.5 K & 1.3 K & Open\\
    MAD~\cite{soldan2022mad}~\footnotemark & 1207.3 h & 110.77 min & 4.1 s & 384.6 K  & 12.7 & 35.5 K & 13.1 K & 5.3 K & Movie\\
    QVHighlights~\cite{lei2021detecting} & 425 h & 2.5 min & 24.6 s & 10.3 K & 11.3  & 4.7 K& 1.8 K& 1.3 K& Vlog / News \\
    Ego4D NLQ~\cite{xu2022ego4d} & 221 & 8.25 min & 9.7 s & 15.1 K & 7.5 & 2.1 K & 0.7 K & 0.4 K & Egocentric\\
    QuerYD~\cite{oncescu2021querydvideodatasethighquality} & 200 h & 4.6 min & 7.7 s & 13.0 K & 19.9 & 10.7 & 5.3 & 3.6 & Open\\
    HiREST~\cite{Zala2023HiREST} & 248 h & 4.4 min & 18.9 s & 8.6 K & 4.4 & 1.8 K &0.5 K &0.4 K & Open \\
    \midrule
    OmniVTG (Ours) & 2124.1 h & 2.76 min & 10.7 s & 359.2 K & 18.9 & 31.2 K & 12.6 K & 22.5 K &  Open \\
    \bottomrule
    \end{tabular}}
    \caption{Statistics of video temporal grounding datasets.}
    \label{tab:dataset_cmp}
\end{table*}

\footnotetext{The videos in MAD dataset are not released, and only the pre-extracted features are available.}

\section{Dataset Collection and Analysis}

We introduce OmniVTG, a large-scale, semantic-rich dataset for video temporal grounding, with 2124 hours of videos and 359,221 text queries and timestamp annotations. Our collection is guided by a novel Semantic Coverage Iterative Expansion pipeline to ensure both massive scale and diverse concept coverage. We first introduce our data collection pipeline and then provide a comprehensive analysis of our dataset in comparison to prior datasets.

\subsection{Semantic Coverage Iterative Expansion}

Our data collection is designed to systematically identify and fill the vocabulary gaps present in existing datasets. To achieve this, we propose a \textit{Semantic Coverage Iterative Expansion} strategy. This process consists of three main stages: target concept identification, iterative targeted video collection, and automated annotation, as shown in Fig.~\ref{fig:pipeline} (a).

\textbf{Target Words Identification.} To define a comprehensive set of target words, we employ the vocabulary from the BERT~\cite{devlin2018bert} tokenizer, which represents a broad spectrum of words used in language. We first clean this vocabulary by performing spelling checks using the Spell Checker library. This approach ensures a broad vocabulary coverage while simultaneously avoiding the inclusion of overly obscure or rare words that have little practical use. We then compute the vocabulary of existing mainstream VTG datasets~\cite{krishna2017dense,gao2017tall,lei2021detecting} and yield a large set of uncovered words that are currently underrepresented. 

\textbf{Interactive Targeted Video Collection.} The purpose of this step is to specifically gather videos that are more likely to contain the target word. For a given target word, directly using it as a search query on online video platforms is often inefficient. For example, searching for a word like `candle' may return many videos where the candle appears in the entire video, which lacks a specific, groundable temporal event. For an abstract and rare concept like `meticulous', a direct search might return irrelevant content. To find videos with a high probability of containing distinct, locatable events, we leverage a powerful LLM (Gemini-2.5 Pro~\cite{comanici2025gemini25pushingfrontier}). We prompt the LLM to generate more specific, event-centric search keywords. For instance, `candle' is translated to `birthday vlog' and `meticulous' is translated to `watchmaker meticulous assembly movement'. These targeted keywords are then used to search relevant videos from online video platforms~\footnote{\url{https://youtube.com/}, \url{https://www.bilibili.com/}}. Then, our Automated Annotation Pipeline will generate timestamps and text queries for the collected videos, ensuring that the target vocabulary is used in the queries as much as possible (specific details are provided below). To iteratively enhance the semantic richness of the dataset, we will further examine the semantic coverage of the collected data after completing a batch of data collection, identify any vocabulary that has not yet been covered, and repeat the above process iteratively. 

\textbf{Automated Annotation.} To annotate the retrieved videos at scale,  we observed that the accuracy of timestamps generated during dense captioning is significantly higher than that obtained through direct grounding for modern MLLMs (Gemini-2.5-Pro~\cite{comanici2025gemini25pushingfrontier}) as shown in Fig.~\ref{fig:teaser_cmp} (c). Based on this insight, we therefore reformulate the task: instead of asking the MLLM to localize a query, we prompt it to describe the events in the video with precise timestamps. Specifically, the MLLM is prompted to generate multiple timestamped captions, with an explicit encouragement to use the target rare words in its descriptions.

\textbf{Test Set Annotation.} Through the above pipeline, we collected a total of 2,124 hours of 46,176 videos, along with 359,221 text queries and corresponding timestamps. To verify the quality of the automated annotations and to obtain a high-quality test set, we randomly selected 10,871 videos for manual validation and correction. We require human annotators to fix any boundary inaccuracies or description errors. This manually-verified subset serves as the official OmniVTG test set for evaluating open-world grounding performance. As shown in Fig.~\ref{fig:teaser_cmp} (c), we compare the accuracy of timestamps before and after manual modifications and find that 93.82\% of the automated labeled timestamps have an IoU greater than 0.5 with the results after manual correction, which demonstrates the quality of our dataset.

\subsection{Dataset Analysis and Comparison}

We now analyze OmniVTG and compare it to existing VTG datasets, with results summarized in Tab.~\ref{tab:dataset_cmp} and  Fig.~\ref{fig:teaser}.

\textbf{Dataset Scale.} As shown in Fig.~\ref{fig:teaser} (c) and Tab.~\ref{tab:dataset_cmp}, OmniVTG is significantly larger than all prior datasets. Our final dataset comprises 2124 hours of video, which is more than 4.3 times larger than ActivityNet Captions~\cite{krishna2017dense} (487h) and 5.0 times larger than QVHighlights~\cite{lei2021detecting} (425h). In terms of annotations, our pipeline generated 359k query-moment pairs, far exceeding other benchmarks (e.g., ActivityNet Captions with 72.0 k queries), as detailed in Tab.~\ref{tab:dataset_cmp}. 

\textbf{Query Semantic Diversity.}  Fig.~\ref{fig:teaser} (c) and Tab.~\ref{tab:dataset_cmp} also show a comparison of the unique vocabulary size, broken down by part-of-speech. As shown in Fig.~\ref{fig:teaser} (c), our OmniVTG demonstrates a massive increase in vocabulary diversity. It covers a much broader range of unique nouns, verbs, and adjectives, even surpassing the vocabulary coverage of all the other datasets combined. Furthermore, in the vocabulary obtained from our Target Words Identification process, the OmniVTG dataset achieved 95\% coverage, while existing public datasets, such as ActivityNet Captions, only reached 48\%.

As detailed in Tab.~\ref{tab:dataset_cmp}, the MAD~\cite{soldan2022mad} dataset is the only one with a comparable scale and wide vocabulary. However, MAD's annotations were generated from audio descriptions, resulting in a strong bias towards very short events (avg. 4.1s duration). Furthermore, the videos in the MAD dataset are not released, and only the pre-extracted features provided by the official are available, which restricts the scope of supported research.

\textbf{Video Domain Diversity.} To assess the domain breadth of the videos in our dataset, we analyze the search keywords used for our targeted video collection. We employed the LLM to categorize these keywords into the domains shown in Fig.~\ref{fig:teaser}(b). This analysis reveals that, unlike domain-specific datasets (e.g., Charades-STA on indoor activities), OmniVTG covers a wide spectrum of topics, further validating its utility for open-world research.

\section{Method}
\label{sec:method}

\subsection{Time Reasoning in VTG Revisited}

Recent work, such as Time-R1~\cite{wang2025timer1}, has successfully applied reinforcement learning to TVG by training the model to generate a chain-of-thought (CoT) before its final answer. Their approach takes a video and a query as input. The model is prompted to first generate a chain-of-thought (CoT) reasoning process, and then provide the final timestamp answer. This output is required to follow a specific format, such as \texttt{<think>...</think> <answer><t\_s to t\_e></answer>}. The framework then verifies the accuracy of the answer and the correctness of the format, using this feedback to optimize the model's policy via Group Relative Policy Optimization (GRPO)~\cite{shao2024deepseekmath}. The composite reward function $r(o)$ used for optimization consists of two main components:
$$r(o) = r_{tIoU}(o) + r_{format}(o)$$
The first component, $r_{tIoU}$, is a timestamp-aware IoU reward that penalizes deviations from the ground-truth start ($t'_s$) and end ($t'_e$) times relative to the total video duration $t$.
$$r_{tIoU}(o) = IoU \cdot (1 - \frac{|t_s - t'_s|}{t}) \cdot (1 - \frac{|t_e - t'_e|}{t})$$
The second component, $r_{format}$, is a binary reward that encourages the model to generate its response in the required reasoning template format.

However, we observe that the reasoning process in Time-R1 lacks explicit guidance, causing the thinking to often default to describing general video or query content, without explicit prediction validation or correction. As a result, its performance in more challenging open-world temporal grounding tasks is suboptimal.

\subsection{Overview}

We observe that MLLMs exhibit stronger capabilities in video understanding (e.g., query-clip matching, event status classification) than in directly grounding, and this robust understanding ability shows a much smaller performance gap between rare and common concepts. Based on this insight, we propose a framework that leverages the model's understanding abilities to perform a self-correction chain-of-thoughts reasoning. We propose a three-stage training paradigm, as shown in Fig.~\ref{fig:pipeline} (b), to achieve this capability: (1) The Supervised Fine-Tuning (SFT) stage enhances the model's basic grounding ability and the specific understanding capabilities that are required during the self-correction process. (2) Self-Correction CoT Finetuning stage establishes the fundamental `predict then correct' chain-of-thought reasoning process. (3) Reinforcement Learning stage further refines and strengthens this reasoning ability on complex examples.

\subsection{Supervised Finetuning}

The goal of this SFT phase is to (1) enhance the model's understanding and grounding abilities regarding the concept of open worlds and (2) explicitly enhance the specific understanding capabilities that are required during the self-correction process. To achieve these, we use a multi-task SFT phase that trains the model on data from OmniVTG. The SFT phase consists of the following four tasks:

\begin{itemize} 
\item \textbf{Temporal Grounding:} Given a text query $Q$, the model predicts the start and end timestamps $[t_s, t_e]$. This enhances the model's basic temporal grounding ability.
\item \textbf{Event Captioning:} Given a time interval $[t_s, t_e]$, the model generates a textual description of the event. This enhances the model's understanding of timestamps and diverse open-world events.
\item \textbf{Query-Clip Matching:} Given a query $Q$ and an interval $[t_s, t_e]$, the model outputs a decision from three categories: match (IoU $> 0.7$), partial match ($0.3 \le$ IoU $\le 0.7$), and mismatch (IoU $< 0.3$). This trains the model to verify its own (or given) predictions.
\item \textbf{Event Status Classification:} Given $Q$ and a specific timestamp $t$, the model predicts the event's status from three options: Not Started, In Progress, Ended. This ability serves the correction phase, as the event status at a specific timestamp can guide the adjustment of predictions. For example, the start time needs to be moved backward if the event has not yet begun at the predicted start time.
\end{itemize}

As shown in Fig~\ref{fig:teaser_cmp}(b), while SFT improves overall performance, we still observe that the grounding performance gap between rare and common concepts persists the video understanding performance is more stable. This supports our further research on how to leverage the model's understanding ability through Self-Correction CoT to improve its grounding performance.

\subsection{Self-Correction Reasoning via CoT Finetuning}
This stage teaches the model to use the skills from SFT to construct a `predict then correct' reasoning path. We reformat the data in our OmniVTG dataset into an explicit CoT template that mimics this process. For a query $B$ from $t_s^B$ to $t_e^B$. We first look for a negative event $A$ from $t_s^A$ to $t_e^A$ in the annotations of the same video. We then reformat these pairs into an explicit CoT reasoning path, which first predicts a coarse prediction $A$ and then corrects it to $B$.
A meaningful negative event $A$ should have visual content similarity with the target event $B$. Therefore, we require that the target event $B$ is temporally enclosed within the event $A$, i.e. $t_s^A \leq t_s^B \quad \text{and} \quad t_e^B \leq t_e^A$. 
For the example above, the CoT data would be: \texttt{I find that $A$ from $t_s^A$ to $t_e^A$. Zooming in further, the event $B$ occurs from $t_s^B$ to $t_e^B$}. 
Finetuning the MLLM on this CoT data explicitly teaches the model to use its video understanding abilities to verify and correct its own localization predictions, and these understanding abilities are more robust in open-world scenarios.

\subsection{Reinforcement Learning for Reasoning}
Finally, to further enhance the model's reasoning abilities on more complex examples, we employ a reinforcement learning stage. We follow the setup of Time-R1~\cite{wang2025timer1} sampling challenging examples (average IoU ~0.3) from our training data and utilizing the GRPO~\cite{shao2024deepseekmath} algorithm for policy optimization. This stage strengthens the `time-aware reasoning' structure established during CoT finetuning, ensuring that the model both maintains the self-correcting, time-aware CoT format and encourages the model to explore and reinforce the content it should first predict in the `predict-then-correct' reasoning path.

\begin{table*}[t]
\centering
\caption{Zero-shot performance comparison with multimodal large language models on video temporal grounding benchmarks.}
\label{tab:public_cmp}
\resizebox{0.9\textwidth}{!}{
\begin{tabular}{lcccccccccccc} 
\toprule
Method & \multicolumn{3}{c}{Charades-STA~\cite{gao2017tall}} & \multicolumn{3}{c}{ActivityNet~\cite{krishna2017dense}} & \multicolumn{3}{c}{QVHighlights~\cite{lei2021detecting}} & \multicolumn{3}{c}{TVGBench~\cite{wang2025timer1}} \\
\cmidrule(lr){2-4} \cmidrule(lr){5-7} \cmidrule(lr){8-10} \cmidrule(lr){11-13} 
 & R1@0.3 & R1@0.5 & R1@0.7 & R1@0.3 & R1@0.5 & R1@0.7 & R1@0.3 & R1@0.5 & R1@0.7 & R1@0.3 & R1@0.5 & R1@0.7 \\
\midrule
ChatVTG~\cite{qu2024chatvtg} & 52.7 & 33.0 & 15.9 & 40.7 & 22.5 & 9.4 &- &- &- & - & - & - \\
TimeChat~\cite{ren2024timechat} & - & 32.2 & 13.4 & 36.2 & 20.2 & 9.5 &- &8.32&  4.26& 22.4 & 11.9 & 5.3 \\
HawkEye~\cite{wang2024hawkeye} & 50.6 & 31.4 & 14.5 & 49.1 & 29.3 & 10.7 & -&- &- & - & - & - \\
VTimeLLM~\cite{huang2024vtimellm} & 51.0 & 27.5 & 11.4 & 44.0 & 27.8 & 14.3 &- &26.1 &11.1 & - & - & - \\
TimeSuite~\cite{zengtimesuite} & 69.9 & 48.7 & 24.0 & -& 16.6 & 9.28 & -& 12.3& 9.16 & 31.1 & 18.0 & 8.9 \\
VideoChat-Flash~\cite{li2024videochat} & 74.5 & 53.1 & 27.6 & - & - & - &- &- &- & 32.8 & 19.8 & 10.4 \\
TRACE~\cite{guo2024trace} & - & 40.3 & 19.4 & - & - & - & -&- &- & 37.0 & 25.5 & 14.6 \\
UniTime~\cite{unitime2025} & -& 59.1 & 31.9 & -& 22.8 & 14.1 & - & 41.0 & 31.5 & - & - & - \\
Time-R1~\cite{wang2025timer1} & 78.1 & 60.8 & 35.3 & 58.6 & 39.0 & 21.4 & 80.3& 66.2& 44.8& 41.8& 29.4 & 16.4 \\
Qwen2.5-VL-7B~\cite{bai2025qwen25vltechnicalreport} & 72.5& 53.6& 28.5& 24.4& 13.6& 6.7& 15.9& 7.10& 4.19& 35.3& 20.0&12.5 \\
\midrule
OmniVTG (Ours) & \textbf{78.3} & \textbf{63.2} & \textbf{37.0} & \textbf{60.3}& \textbf{39.8}& \textbf{21.4}& \textbf{82.8} & \textbf{67.0} & \textbf{47.3} &\textbf{ 54.5}  &\textbf{ 37.6} & \textbf{19.7} \\
\bottomrule
\end{tabular}
}
\end{table*}

\begin{table*}[t]
\centering
\caption{Zero-shot performance comparison on open-word video temporal grounding when queries contain rare concepts.}
\label{tab:omnivtg_cmp}
\resizebox{0.9\textwidth}{!}{%
\begin{tabular}{l|cccccccccccc}
\toprule
\multirow{3}{*}{Method} & \multicolumn{6}{c}{OmniVTG Test Set (Ours)} & \multicolumn{6}{c}{ActivityNet Captions~\cite{krishna2017dense}}  \\
\cmidrule(lr){2-7} \cmidrule(lr){8-13}
& \multicolumn{3}{c}{Full} & \multicolumn{3}{c}{Rare} & \multicolumn{3}{c}{Full} & \multicolumn{3}{c}{Rare} \\
\cmidrule(lr){2-4} \cmidrule(lr){5-7} \cmidrule(lr){8-10} \cmidrule(lr){11-13}
& R1@0.3 & R1@0.5 & R1@0.7 & R1@0.3 & R1@0.5 & R1@0.7 & R1@0.3 & R1@0.5 & R1@0.7 & R1@0.3 & R1@0.5 & R1@0.7 \\
\midrule
UniTime~\cite{unitime2025}~\tablefootnote{UniTime only released the model fine-tuned on ActivityNet Captions, so we are unable to report its zero-shot performance on Rare set. The performance on the Full set is cited from the original paper.} & 59.9 & 37.1 & 15.8 & 54.2 & 34.9& 12.7& 39.9&22.8 & 14.1& - &- & - \\
Time-R1~\cite{wang2025timer1} & 57.1 & 33.9 & 14.7 & 49.7 & 29.4& 15.7& 58.6& 39.0& 21.4& 56.2 &36.1 & 19.3 \\
Qwen2.5-VL-7B~\cite{bai2025qwen25vltechnicalreport} &49.0 &34.5 &16.9 & 44.7 & 31.5 &15.7 & 24.4 &13.6 &6.70 & 22.3 & 12.9 & 4.8 \\
\midrule
OmniVTG (Ours) & \textbf{74.2}&\textbf{63.3} &\textbf{47.6} & \textbf{74.1}& \textbf{62.4}& \textbf{46.2}& \textbf{60.3}& \textbf{39.8}& \textbf{21.4}&\textbf{60.1} &\textbf{39.5} & \textbf{20.8}\\
\bottomrule
\end{tabular}%
}
\end{table*}

\begin{table}[t]
\centering
\caption{Ablation studies on the OmniVTG test set and ActivityNet Captions. We report the performance on the metric R1@0.5.}
\label{tab:ablation}
\resizebox{0.9\linewidth}{!}{%
\begin{tabular}{l|ccc}
\toprule
Model & \makecell{OmniVTG (Full)} & \makecell{OmniVTG (Rare)} & \makecell{ActivityNet} \\
\midrule
\multicolumn{4}{l}{\textit{1. Necessity of Training Stages}} \\
Qwen2.5-VL-7B & 34.5 & 31.5& 13.6 \\
+ SFT & 62.3 & 59.6 & 25.6\\
+ SFT + COT & 62.4 & 61.3 & 32.5\\
+ SFT + RL & 62.8 & 60.6 & 37.2 \\
+ SFT + COT + RL & \textbf{63.3} & \textbf{62.4}& \textbf{39.8}\\
\midrule
\multicolumn{4}{l}{\textit{2. Impact of SFT Data Scale}} \\
SFT (10\% data) & 41.9& 37.8 & 15.3 \\
SFT (50\% data) & 58.7 & 55.4 & 21.9 \\
SFT (100\% data) &\textbf{62.3} &\textbf{59.6} & \textbf{25.6}\\
\midrule
\multicolumn{4}{l}{\textit{3. Comparison of Reasoning Strategy}} \\
w/o Reasoning & 62.3 & 59.6 & 25.6 \\
Rule-base reflection &  62.4 & 61.0 & 37.9 \\
Content-aware reflection & \textbf{63.3} & \textbf{62.4} & \textbf{39.8} \\
\bottomrule
\end{tabular}
}
\end{table}

\section{Experiments}
\label{sec:experiments}

\subsection{Datasets and Evaluation}
\label{sec:datasets}

\noindent\textbf{Zero-Shot Evaluation Benchmarks.}
To assess the generalization and open-world capabilities of our model, we evaluate its zero-shot performance on four established public benchmarks: ActivityNet Captions~\cite{krishna2017dense} (human activities), Charades-STA~\cite{gao2017tall} (indoor activities), QVHighlights~\cite{lei2021detecting} (Vlogs and news), and TVGBench~\cite{wang2025timer1} (a comprehensive benchmark designed to evaluate the temporal grounding capabilities across diverse query types).

\noindent\textbf{Rare Concept Evaluation.}
To fully understand the model's open-world grounding capabilities, especially on rare concepts, we evaluate on our manually-corrected OmniVTG test set. In addition to testing on the complete test set, we also divided a subset containing rare concepts to understand the performance gap between rare and non-rare concepts~\footnote{We define rare concepts using the \texttt{wordfreq} library. Any word in the query with a frequency of less than 1e-7 in the library's reference corpus is classified as rare.}. Given that our OmniVTG test set has the same distribution as the training data, which may lead to an unfair comparison, we also divide a Rare subset from the unseen ActivityNet Captions~\cite{krishna2017dense} dataset for further evaluation.

\noindent\textbf{Evaluation Metrics.}
Following standard practice~\cite{wang2025timer1,unitime2025}, we report temporal grounding performance using Recall@1 (R1) at various Intersection-over-Union (IoU) thresholds.

\subsection{Implementation Details}
\label{sec:implementation}

We conduct our experiments on the Qwen2.5-VL-7B~\cite{bai2025qwen25vltechnicalreport} baseline. 
For the SFT and CoT finetuning stages, we train the model using LoRA (rank=8, $\alpha=8$) with a learning rate of 2e-4. For the Reinforcement Learning (RL) stage, we employ the GRPO algorithm to perform full-parameter fine-tuning of the LLM with a learning rate of 1e-6. Throughout all stages, the vision encoder remains frozen. 

\subsection{Comparison with State-of-the-Art Methods}
\label{sec:sota}

We compare our model with other MLLMs for video temporal grounding in the zero-shot setting on public benchmarks, then evaluate top performers, UniTime~\cite{unitime2025} and Time-R1~\cite{wang2025timer1}, on the OmniVTG dataset to assess open-world performance, particularly with rare concepts.

\textbf{Zero-Shot Performance on Public Benchmarks.}
As shown in Tab.~\ref{tab:public_cmp}, our model, trained only on OmniVTG, achieves state-of-the-art zero-shot performance across all four public benchmarks. Notably, the base Qwen2.5-VL-7B model performs poorly, achieving only 13.6\% R1@0.5 on ActivityNet and 7.10\% on QVHighlights. Our model (OmniVTG) significantly improves the base model's performance and outperforms other MLLMs designed for video temporal grounding. Compared to the strong Time-R1 baseline, our model shows consistent improvements, especially on the TVGBench, which contains diverse data sources and question types. As we can see, our model achieves 52.8\% on R1@0.3, substantially outperforming Time-R1 (41.8\%) and TRACE (37.0\%). This demonstrates our superior generalization capability.

\textbf{Open-World Temporal Grounding Performance.}
We further analyze performance on our open-world video temporal grounding dataset OmniVTG and compare the performance when the query contains rare concepts in Tab~\ref{tab:omnivtg_cmp}. On our OmniVTG Test Set, our model achieves 63.3 R1@0.5, significantly outperforming Time-R1 (33.9\%) and UniTime (37.1\%). More importantly, on the OmniVTG Rare subset, our model's performance is close to its performance on the full set. In contrast, Time-R1's performance drops from 33.9\% to 29.4\%.
This trend is confirmed on the ActivityNet dataset. On the ActivityNet Rare subset, our model's performance remains stable, while Time-R1's performance drops from 39.0\% to 36.1\%. This demonstrates that our method is not only more accurate overall but also more robust to rare concepts, closing the performance gap between rare and common concepts in the open-world scenario.

\subsection{Ablation Studies}
\label{sec:ablations}

Tab.~\ref{tab:ablation} shows the ablation studies on the OmniVTG test set and ActivityNet Captions to validate our design choices.

\textbf{Necessity of Training Stages.}
Part 1 of Tab.~\ref{tab:ablation} shows the contribution of each training stage. The base Qwen2.5-VL-7B model performs poorly. Adding our multi-task SFT stage provides the most significant boost, not only improving performance on the OmniVTG test set, but also significantly enhancing performance on the unseen ActivityNet Captions dataset (from 13.6\% to 25.6\%). Adding COT on top of SFT further improves performance, especially on the Rare subset (from 59.6\% to 61.3\%) and ActivityNet (from 25.6\% to 32.5\%), validating that our self-correction reasoning improves generalization. The full model, SFT + COT + RL, achieves the best performance, demonstrating that all three stages are complementary and essential. Notably, when removing the COT finetuning and directly applying reinforcement learning (SFT+RL), the performance will drop, especially on the OmniVTG Rare set and ActivityNet Captions datasets. This proves that our explicit self-correction CoT provides a superior reasoning path.

\textbf{Impact of SFT Data Scale.}
In Part 2 of Tab.~\ref{tab:ablation}, we analyze the impact of our SFT data scale. Training with only 10\% of data yields 41.9\% R1@0.5 on the OmniVTG test set. This performance scales consistently as data increases, from 58.7\% with 50\% data to 62.3\% with 100\% data. This trend is mirrored on the ActivityNet benchmark,  confirming the effectiveness of our large-scale dataset.

\textbf{Comparison of Reasoning Strategy.}
In Part 3 of Tab.~\ref{tab:ablation}, we compare two reasoning strategies: the rule-based and content-based self-correction COT. The rule-based method randomly shifts the ground truth boundaries for the initial localization, while the content-based method locates segments in the video similar to the ground truth. Both strategies outperform the baseline without reasoning, highlighting the importance of explicit self-reflection. Among them, the Content-aware reflection performs best, showing that using semantically similar video content for the initial prediction leads to better results.

\begin{figure}
    \centering
    \includegraphics[width=0.9\linewidth]{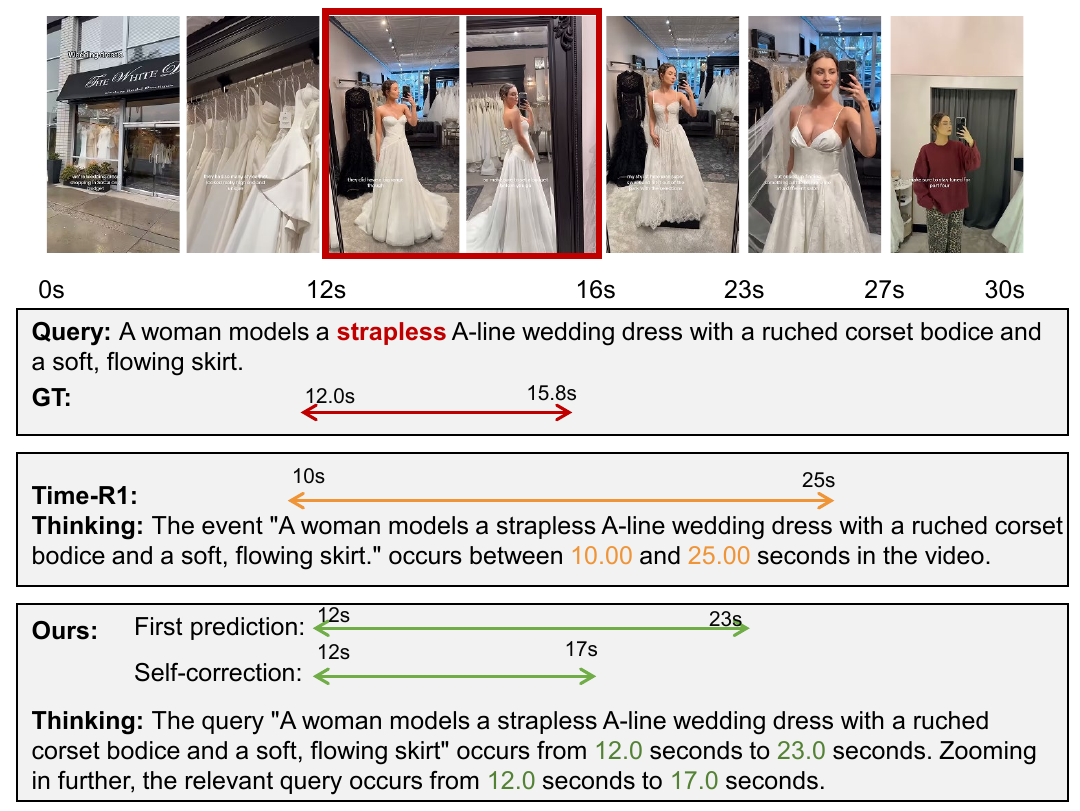}
    \caption{Qualitative comparison with Time-R1.}
    \label{fig:vis}
\end{figure}

\subsection{Qualitative Results}

Fig.~\ref{fig:vis} shows a qualitative comparison between our method and Time-R1. The keyword in the query is `strapless'. Both Time-R1 and our method initially include incorrect segments where the model is trying on another wedding dress with straps. However, our model corrects this mistake during the self-correction phase, yielding the correct prediction, while Time-R1 directly returns the wrong answer.

\section{Conclusion}
\label{sec:conclusion}

This paper introduces OmniVTG, a large-scale dataset for open-world Video Temporal Grounding. By the Semantic Coverage Iterative Expansion pipeline, OmniVTG significantly improves the scale and semantic richness. We also propose a Self-Correction Chain-of-Thought (CoT) training paradigm, which enables MLLMs to refine their predictions through reflection and correction. Our experiments demonstrate that we achieve SOTA performance not only on our OmniVTG dataset but also on four public VTG datasets.

\noindent\textbf{Acknowledgements.} 
This work was supported by the grants from the National Natural Science Foundation of China (62372014, 62525201, 62132001, 62432001), Beijing Nova Program, Beijing Natural Science Foundation (4252040, L247006), and Wuhan East Lake High-Tech Development Zone National Comprehensive Experimental Base for Governance of Intelligent Society.

{
    \small
    \bibliographystyle{ieeenat_fullname}
    \bibliography{main}
}


\end{document}